\documentclass[letterpaper, 10 pt, conference]{ieeeconf}
\IEEEoverridecommandlockouts
\let\labelindent\relax

\usepackage{enumitem}
\usepackage{times}
\usepackage{cite,url}
\usepackage{multicol}
\usepackage{bm}
\usepackage{xcolor}
\usepackage{amsfonts}
\usepackage{booktabs}
\usepackage{balance}
\usepackage{amsmath, amssymb}
\usepackage{graphicx}
\usepackage[font={small}]{caption}
\usepackage{subcaption}
\usepackage[pagebackref=true,breaklinks=true,colorlinks=true,bookmarks=true,citecolor=blue]{hyperref}

\usepackage{amsthm}

\title{\LARGE \bf Adapting Rapid Motor Adaptation for Bipedal Robots}
\author{Ashish Kumar*$^{1}$, Zhongyu Li*$^{1}$,  Jun Zeng$^{1}$, Deepak Pathak$^{2}$, Koushil Sreenath$^{1}$, Jitendra Malik$^{1}$ 
\thanks{$^*$ Authors contributed equally.} 
\thanks{$^{1}$ University of California Berkeley, \{ashish\_kumar, zhongyu\_li, zengjunsjtu, koushils, malik\}@berkeley.edu}%
\thanks{$^{2}$ Carnegie Mellon University, dpathak@cs.cmu.edu }%
}

\begin{document}
\maketitle

\begin{abstract}
Recent advances in legged locomotion have enabled quadrupeds to walk on challenging terrains. However, bipedal robots are inherently more unstable and hence it's harder to design walking controllers for them. In this work, we leverage recent advances in rapid adaptation for locomotion control, and extend them to work on bipedal robots. Similar to existing works, we start with a base policy which produces actions while taking as input an estimated extrinsics vector from an adaptation module. This extrinsics vector contains information about the environment and enables the walking controller to rapidly adapt online. However, the extrinsics estimator could be imperfect, which might lead to poor performance of the base policy which expects a perfect estimator. In this paper, we propose A-RMA (Adapting RMA), which additionally adapts the base policy for the imperfect extrinsics estimator by finetuning it using model-free RL. We demonstrate that A-RMA outperforms a number of RL-based baseline controllers and model-based controllers in simulation, and show zero-shot deployment of a single A-RMA policy to enable a bipedal robot, Cassie, to walk in a variety of different scenarios in the real world beyond what it has seen during training. Videos and results at~\url{https://ashish-kmr.github.io/a-rma/}
\end{abstract}

\IEEEpeerreviewmaketitle


\section{Introduction}
\label{sec:intro}
The ability of legged animals to traverse a wide variety of terrains has inspired decades of research in legged robots to replicate this capability in artificial systems. Although great advances have been made via control theoretic methods
~\cite{vukobratovic2004zero,kuindersma2014efficiently,pratt2012capturability,xiong2018coupling,feng20133d,dai2014whole,GrChAmSi2010,radford2015valkyrie,da20162d,gong2020angular,gong2019feedback,li2020animated,reher2021control,dantec2021whole}, designing these controllers often requires expert parameter tuning for different terrain. With recent advances in learning, research interest has shifted towards methods that \textit{learn} to walk. Impressive results have been achieved in the past few years, notably Lee et.al.~\cite{lee2020learning} and RMA~\cite{rma}, for general-purpose legged locomotion in challenging real-world terrains. However, these results have largely been shown in quadrupeds which are relatively easier to control than bipedal robots.

\begin{figure}
    \centering
    \includegraphics[width=\linewidth]{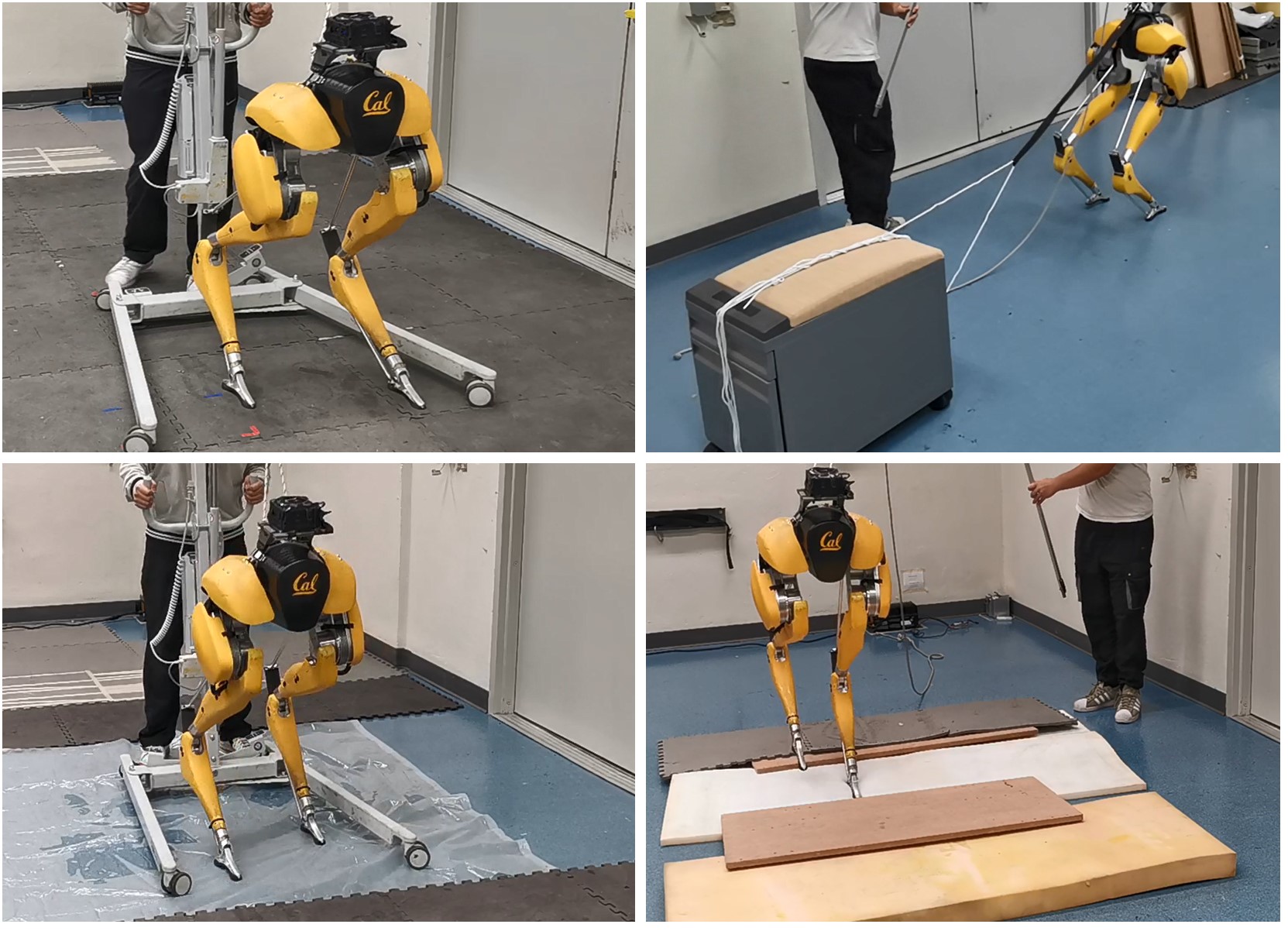}
    \caption{We demonstrate the performance of A-RMA with a bipedal robot in several challenging setups which includes slippery surfaces, foam, and pulling a payload. Some of these were never seen during training and the policy was deployed without any calibration or real-world finetuning. Note that bipeds are inherently more unstable compared to quadrupeds, which makes each of these much more challenging than for a quadruped robot. Videos at~\url{https://ashish-kmr.github.io/a-rma/}}
    \label{fig:main-figure}
\end{figure}

Our goal is to see how well a learning-based approach works for bipedal robots. Unlike their quadruped or hexapod counterparts, bipeds are more dynamic and offer coverage of more terrains at the cost of being more prone to instability. Although there have been several investigations of applying learning methods to bipedal robots~\cite{li2021reinforcement, siekmann2020learning}, until now, the robustness performance has not matched their quadruped counterparts in terms of robustness to scenarios not seen during training (Figure~\ref{fig:main-figure}).

We train our robot in simulation and start with Rapid Motor Adaptation~(RMA)~\cite{rma} as a baseline for sim2real transfer to the Cassie biped. In contrast to the popular idea of training a terrain-invariant policy~\cite{tobin2017domain}, RMA trains an adaptive policy conditioned on latent \textit{extrinsics} vector that encodes terrain-specific information in simulation. After training this base policy, the extrinsic vector is then estimated online by an adaptation module using the history of observed proprioceptive states. However, we found that scaling RMA as-is to complex bipeds is faced with several challenges.

\begin{figure}[t]
      \centering
      \includegraphics[scale=0.18]{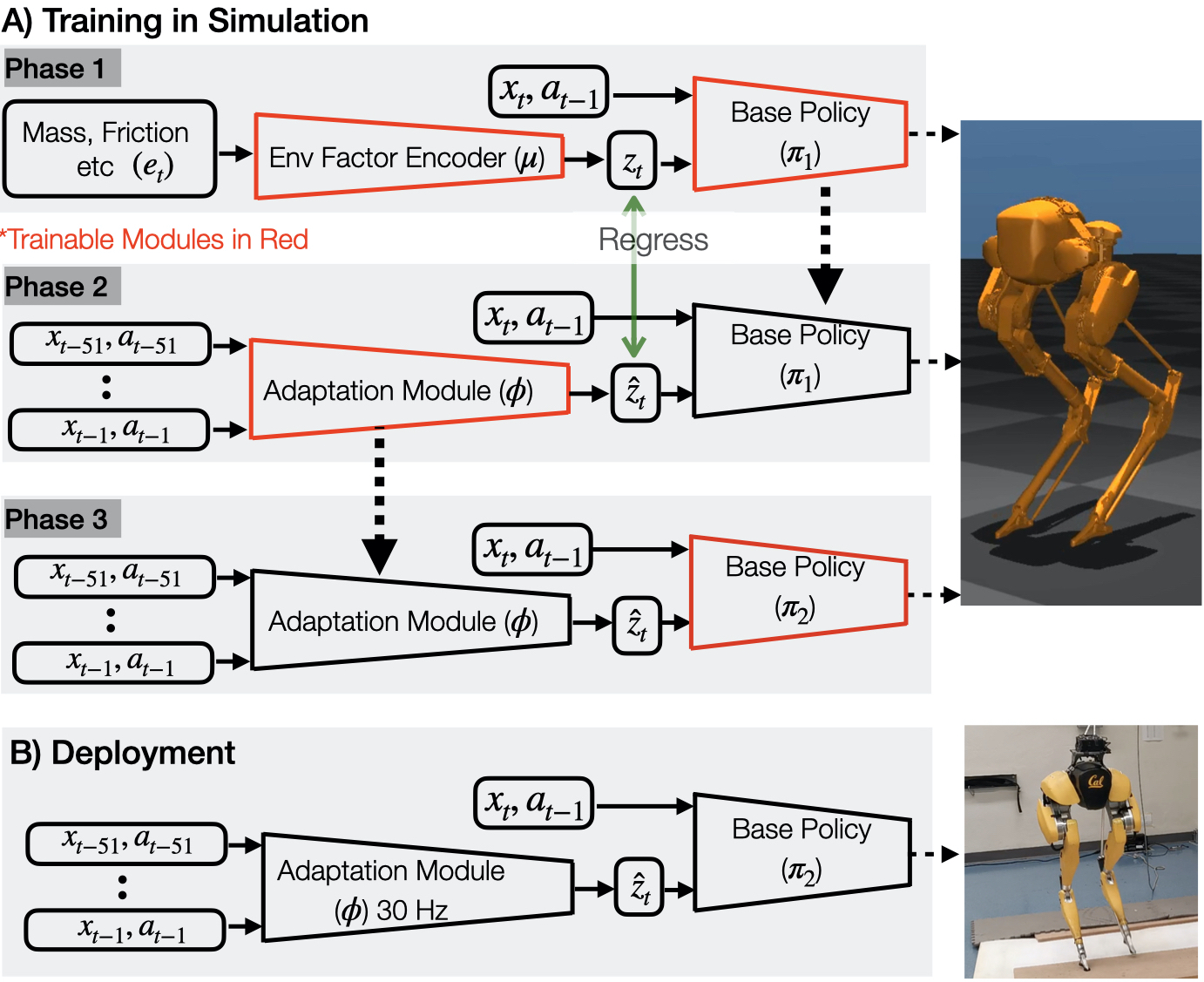}
      \caption{We show the training and deployment phases of A-RMA. The first two phases are the same as RMA~\cite{rma}. We additionally add a third phase in which the base policy is fine-tuned again with PPO while keeping the adaptation module fixed, to account for imperfect estimation of extrinsics. We found this to be critical for reliable performance in the real world.}
      \label{fig:method}
      \vspace{-0.12in}
   \end{figure}

The first challenge is training the base policy from scratch using reinforcement learning. To address this issue and learn naturally stable high-performance gaits, we bootstrap the base policy from a set of reference motions generated using a gait library. The second challenge, however, is a more fundamental one. For complex robots like bipeds, it is usually very challenging to precisely estimate the privileged extrinsics at deployment just from the observable states. This creates a large domain gap for the base policy at deployment which was trained with accurate extrinsics. The higher the magnitude of extrinsics estimation error, the worse the base policy performs. We bridge this gap by another round of finetuning the base policy using the imperfect extrinsics estimated from adaptation module instead of conditioning on the perfect extrinsics. We call our method \textit{A-RMA} due to the adaptation of RMA policy itself, as outlined in Figure~\ref{fig:method}. Similar to the base policy and adaptation module in RMA, A-RMA \textit{policy adaptation} step is also trained in simulation. We evaluate A-RMA on a complex Cassie bipedal robot and show evaluations on several challenging environments in both simulation and real world.


\section{Related Work}
\label{sec:relatedwork}
\paragraph{Model-based Control for Bipedal Robots}
Locomotion of bipedal robots has been traditionally approached via notions of gait stability such as ZMP criterion~\cite{vukobratovic2004zero} or Capturability~\cite{pratt2012capturability} using robot's reduced-order models~\cite{kuindersma2014efficiently,xiong2018coupling,li2021vision,feng20133d,dai2014whole,gong2020angular}. 
Although these methods can effectively control humanoid robots with flat feet, they often walk conservatively. 
Alternatively, Hybrid Zero Dynamics~(HZD)~\cite{GrChAmSi2010,da20162d, gong2019feedback,li2020animated,reher2021control} based techniques can also generate stable periodic walking based on input-output linearization using robot's full-order model. 
However, HZD-based controllers for 3D robots typically need extensive parameter tuning in both simulation and in the real world, and are hard to adapt to the environment changes. 
In this work, we use reinforcement learning to learn our walking controllers, using the HZD-based walking gaits only as reference motions to produce natural looking gaits. 
This allows us to generate more diverse and effective stable walking behaviours since we don't have hard constraints on periodicity and neither do we require a precise model.

\paragraph{RL-based Control for Legged Robots}
Reinforcement learning for legged locomotion has shown promising results in learning walking controllers that can be successfully deployed in the real world~\cite{li2021reinforcement,rma,kohl2004policy,hwangbo2019learning,peng2020learning}. Data-driven methods could either use reference motions~\cite{peng2020learning,li2021reinforcement} to learn walking behaviours, learn residuals over predefined foot motions~\cite{lee2020learning, iscen2018policies}, or learn without any motion priors and foot trajectory generators~\cite{rma,haarnoja2018learning,hwangbo2019learning}. 
In this work, we use a hybrid approach where we warm start the learning process using a HZD-based gait library, but subsequently reduce the dependency on them by lowering the costs related to imitating the gait library. 

Recent works have also built bipedal controllers using RL. 
Model-based RL in~\cite{castillo2020velocity, castillo2020hybrid} is used to learn a walking controller on Cassie in simulation to track a velocity. 
Alternatively, work in ~\cite{xie2018feedback,xie2020learning} learns model free residuals over reference motions, while work in ~\cite{johannink2019residual} learns walking policies that can track a commanded planar velocity on Cassie in the real world. Although residual control speeds up training, the corrections that can be applied to the reference trajectory is limited which limits the diversity of behaviours of the controller. This can be addressed by learning policies which don't rely on reference motions during deployment, but instead add motion constraints during training by either using reference motions or through foot fall constraints via reward terms~\cite{li2021reinforcement,siekmann2021blind,siekmann2021sim}. We follow this general approach in this paper. More specifically, we use reference trajectories for imitation via reward terms, except that we only use them to warm start the learning process and then reduce the dependency on them and use energy minimization~\cite{fu2021minimizing} to allow for diverse behaviors which can be more general than the reference motions. 

\paragraph{Simulation to Real World Transfer}
Sim-to-real methods allow deployment of walking policies in the real world after training them in simulation, which is a safe and inexpensive data source. Parameters finetuning on the hardware is one way to bridge the gap for model-based methods~\cite{gong2019feedback,li2020animated,yang2022bayesian}. Alternatively, several works use domain randomization which varies the system properties in simulation in order to cover the uncertainty in the real world~\cite{sadeghi2016cad2rl,tobin2017domain,peng2018sim,yu2019sim,yu2020learning,peng2020learning,li2021reinforcement,siekmann2020learning,siekmann2021blind,siekmann2021sim}. These methods achieve robustness at the cost of optimality, where they try to learn a single environment-agnostic behaviour for all deployment scenarios. Instead, we learn an adaptive policy which enables transfer to widely varying deployment scenarios.

\paragraph{System Identification and Adaptation}
An adaptive policy conditions to environment variations instead of being agnostic to them. These variations can be explicitly estimated during deployment either through a module that is trained in simulation~\cite{yu2017preparing} or can be estimated by optimizing for high returns using evolutionary algorithms~\cite{yu2018policy}. Predicting the exact parameters is both unnecessary and difficult, which in turn leads to poor empirical performance.
Instead, a low dimensional latent embedding can be used~\cite{rma,peng2020learning,zhou2019environment} which contains an implicit estimate of the environment. At test time, this latent embedding can be optimized using policy gradients from real-world rollouts~\cite{peng2020learning}, Bayesian optimization~\cite{yu2019biped}, or random search~\cite{yu2020learning}. An alternate approach is to use meta learning to learn to adapt online~\cite{finn2017model}. These methods tend to take multiple rollouts to adapt in the real world~\cite{song2020rapidly, clavera2018learning} which is prohibitive for several scenarios. In this work, we follow the approach of~\cite{rma} which learns a feed forward policy to estimate the latent environment vector, enabling rapid adaptation in fractions of a second.

\section{General Walking Controller}
Our walking controller contains a base policy which produces the target joint points for the robot, and an adaptation module which uses proprioceptive history to continually estimate the extrinsics vector.
Such a vector contains information about the environment that can be used by the base policy to adapt. In this section, we describe how we train the base policy for bipeds since trivially training the policy from scratch results in unnatural gaits, unlike prior work in quadrupeds~\cite{rma}. Instead we use reference motions to bootstrap learning but unlike prior methods which use reference motions~\cite{lee2020learning,li2021reinforcement}, we reduce the dependence on these reference motions as training progresses to learn optimal behaviors in a data driven way.

\subsection{Walking Policy} 
We train an adaptive walking policy ($\pi_1$) using model free reinforcement learning to learn a stable walking policy while imitating reference motions generated using a HZD-based gait library. 
Here, the gait library consists of a collection of periodic walking gaits at different walking velocities and walking heights. 
The base walking policy ($\pi$) takes two arguments as input 1) the current state $x_t$, 2) the extrinsics vector $z_t \in \mathbb{R}^{8}$ and then predicts the next action $a_t$. The predicted action $a_t$ is the target position for the $10$ actuated robot joints which is converted to torque using a PD controller. The extrinsics vector $z_t$ is a compressed version of the environment vector $e_t \in \mathbb{R}^{147}$ generated by $\mu$. The environment vector $e_t$ includes things such as friction, inertia, center of mass, etc (see Sec III-C), but the extrinsics vector $z_t$ only retains as much information from $e_t$ as is necessary for the base policy to adapt. We thus have,
\begin{align}
    z_t &= \mu(e_t),  \label{eq:mu}  \\
    a_t &= \pi_1(x_t, a_{t-1}, g_t, z_t). \label{eq:pi}
\end{align}

We implement $\mu$ and $\pi_1$ as MLPs and jointly train them end-to-end using model-free reinforcement learning to maximize the following expected return of the policy $\pi$: 
\[
    J(\pi) = \mathbb{E}_{\tau \sim p(\tau|\pi)}\Bigg[\sum_{t=0}^{T-1}\gamma^t r_t\Bigg],
\]
where $\tau = \{(x_0, a_0, r_0), (x_1, a_1, r_1) . . .\}$ is the trajectory of the agent when executing policy $\pi_1$, and $p(\tau|\pi_1)$ represents the likelihood of the trajectory under $\pi_1$.

\subsection{Reward Function} \label{sec:rew}
The reward function encourages the agent to track the goal commanded and to imitate the reference motion from the gait library. Let's denote the robot's actual motor positions as $\hat{q}_m$, pelvis translation position and velocity as $\hat{q}_p$ and $\hat{\dot{q}}_p$, the pelvis rotation and rotation velocity as $\hat{\dot{q}}_r$ and $\hat{q}_r$. Let the corresponding quantities for the reference motion be denoted by $q_m^r$, $q_p^r$, $\dot{q}_p^r$, $q_r^r$ and $\dot{q}_r^r$ respectively. We additionally denote the robot's torque as $u$ and the robot's ground reaction force as $F$. The reward at each time step $t$ is a weighted sum of:
\begin{enumerate}
    \item Motor imitation: $\exp[-\rho_1||q_m^r - \hat{q}_m||^2_2]$
    \item Pelvis position imitation: $\exp[-\rho_2||q_p^r - \hat{q}_p||^2_2]$
    \item Pelvis velocity imitation: $\exp[-\rho_3||\dot{q}_p^r - \hat{\dot{q}}_p||^2_2]$
    \item Pelvis rotation imitation: $\exp[-\rho_4(1-\cos{(q_r^r - \hat{q}_r)}]$
    \item Rotational velocity imitation: $\exp[-\rho_5||\dot{q}_r^r - \hat{\dot{q}}_r||^2_2]$
    \item Torque penalty: $\exp[-\rho_6||u||^2_2]$
    \item Ground reaction force penalty: $\exp[-\rho_7||F||^2_2]$,
\end{enumerate}
where the corresponding weights of each of the reward terms are $[0.3, 0.24, 0.15, 0.13, 0.06, 0.06, 0.06]^T$, and $\rho_1$ to $\rho_7$ are $[5.0, 0.1, 5.0, 5.0, 1.0, 5e^{-7}, 1.25e^{-5}]$). To stabilize the pelvis, we set the desired roll and pitch velocity to $0$, and the desired yaw velocity is from the user command $c(t)$. We compute the desired pelvis translational and rotational positions by integrating the corresponding desired velocity. Moreover, we decay the coefficients of the imitation terms as the training progresses to reduce the dependence on reference motions and learn optimal data driven behaviours.

\subsection{Environment Variations and Terrains} During the learning of the walking policy, we vary the ground friction, robot's motor friction, mass and center of mass of the robot and robot links. We train over fractal terrain, flat terrain, slopes and discrete terrains such as steps (see Table~\ref{tab:randomization}). 

\section{Walking Controller with Rapid Adaptation}
Following the process described in the previous section, we now have a base policy which can use the extrinsics vector to adapt. 
However, this extrinsics vector uses privileged simulation information and is unavailable in the real world. 
To resolve this, we generalize the idea proposed in~\cite{rma} of learning an adaptation module using the history of commanded actions and robots proprioception to estimate these extrinsics online. 
The insight is that the discrepancy between what was commanded to the robot and the actual joint positions contains information about these extrinsics. 
However, such an estimation could be noisy as the extrinsics vector might not be fully observable. 
To fix this, we further finetune the base policy to learn to walk adaptively with an imperfect extrinsics vector. We now describe this in detail below.

\subsection{Adaptation Module for Estimating Extrinsics}
Privileged environment information $e_t$ is not available in the real world, and consequently its encoded extrinsics vector $z_t$ is not accessible during deployment. Similar to ~\cite{rma}, we estimate the extrinsics online from the proprioceptive history ($x_{t-k:t-1}$, $a_{t-k:t-1}$) using the adaptation module $\phi$. In our experiments, we use $k = 70$ which roughly corresponds to $2$s. The estimated extrinsic vector is thus given by,
$$\hat{z_t} =  \phi\big(x_{t-k:t-1}, a_{t-k:t-1}\big).$$

We can train the adaptation module in simulation via supervised learning to minimize: $\text{MSE}(\hat{z_t}, z_t) = \| \hat{z_t} - z_t\|^2,$
where $z_t = \mu(e_t)$. We model $\phi$ as a $1$-D CNN to capture temporal correlations. 

We collect trajectories by unrolling the base policy $\pi_1$ with the $\hat{z_t}$ predicted by the randomly initialized function $\phi$, and then pair it with the ground truth $z_t$ to train $\phi$. We iteratively repeat this until convergence.  

\subsection{Finetuning with Estimated Extrinsics}
For our setting, the extrinsics estimated by the adaptation module $\phi$ is imperfect since $z_t$ might not be fully observable from the proprioception history. We observed this as the regression error to estimate $z_t$ did not become low enough during the training of the adaptation module. This causes the base policy $\pi_1$ to experience a significant drop in performance since it is trained with perfect extrinsics, and noisy estimation introduces a domain gap. To overcome this issue, we propose to further finetune the base policy $\pi_1$ with the imperfect extrinsics predicted by the adaptation module $\phi$ trained in phase 2, see Fig.~\ref{fig:method}. The adaptation module is kept frozen and the base policy is finetuned using model free reinforcement learning with the reward described in Section~\ref{sec:rew}. This gives us the final base policy $\pi_2$ which is used in deployment.


\section{Experimental Setup}

\subsection{Hardware} We use the Cassie robot for our experiments. It is a dynamic, life-sized, and underactuated bipedal robot which has $20$ DoFs and is introduced in details in~\cite[Sec. II]{li2020animated}. It has 10 actuated rotational joints $q_m = [q_{1,2,3,4,7}^{L/R}]^T$ (abduction, rotation, hip pitch, knee, and toe motors), and four passive joints $q_{5,6}^{L/R}$ (shin and tarsus joints). 
Its floating base, the robot pelvis $q_p = [q_{x}, q_{y}, q_{z}, q_{\psi}, q_{\theta}, q_{\phi}]^T$, has 6 DoFs for sagittal $q_x$, lateral $q_y$, vertical $q_z$, roll $q_{\psi}$, pitch $q_{\theta}$, and yaw $q_{\phi}$, respectively.
The entire robot coordinate $q \in \mathbb{R}^{20}$ includes all robot DoFs whereas the observable DoFs $q^o \in \mathbb{R}^{17}$ excludes the pelvis translational position $q_{x,y,z}$ which can not be reliably obtained by the sensors on the robot hardware. 
There is an onboard Intel NUC on the robot, which runs our policy. The policy updates at $30$~Hz and the joint-level PD controller operates at $2$~kHz.

\subsection{Simulation Setup} We use the simulation environment for reinforcement learning on Cassie developed in~\cite{li2021reinforcement}, which itself is based on an open source MuJoCo simulator\cite{todorov2012mujoco,cassiesim}. 
During training, each episode has a maximum number of time steps of  $T$=$2500$ ($83$~s). 
In each episode, we resample a new command $c(t)=[\dot{q}^d_x~\dot{q}^d_y~q^d_z, \dot{q}^d_{\phi}]^T$ and randomize the environment parameters every $8$~s. 
The randomization range of the command is between $[-1.0, -0.3, 0.65, -30^{\circ}]^T$ and $[1.0, 0.3, 1.0, 30^{\circ}]^T$. 
The episode will terminate early if the pelvis height falls below $0.55$~m, or if the tarsus joints $q^{L/R}_6$ hit the ground~\cite{li2021reinforcement}. 

\begin{table}[t]
\centering
\begin{tabular}{ccc}
\hline
\textbf{Parameter}         & \textbf{Range}            & \textbf{Unit}      \\ \hline
Link Mass                  & {[}0.7,1.3{]} $\times$ default & kg        \\
Link Mass Center           & {[}0.7,1.3{]} $\times$ default & m         \\
Joint Damping              & {[}0.3,4.0{]} $\times$ default & Nms/rad   \\
Spring Stiffness              & {[}0.95,1.05{]} $\times$ default & Nm/rad   \\
Ground Friction Ratio      & {[}0.3, 3.0{]}            & 1                \\
Fractal Terrain Height      & {[}0.0, 0.12{]}              & m               \\ \hline
\end{tabular}
\caption{The range of dynamics parameters that are extended from~\cite[Table II]{li2021reinforcement} and are varied in the phase 1 of A-RMA.}
\label{tab:randomization}
\vspace{-2mm}
\end{table}

\subsection{State-Action Space}
\paragraph{Observation} 
The observation $\mathbf{x}_t = (\mathbf{q}^o_{t-4:t}, \mathbf{a}_{t-4:t-1}, \mathbf{q}^r_{m,t},\mathbf{q}^r_{m,t+1},\mathbf{q}^r_{m,t+4},\mathbf{q}^r_{m,t+7}, c_t)$ at time $t$ consists of 4 parts.
There are observable robot states $\mathbf{q}^o=[q^o,\dot{q}^o]$ at the current and the past 4 time steps, and the actions $\mathbf{a}$ from the past 4 time steps.
There is also reference motion $\mathbf{q}^r_m=[q^r_m,\dot{q}^r_m]$ which includes the reference motor position and motor velocity at current time $t$ and future time steps (next 1, 4, 7 time steps) obtained from a HZD-based gait library~\cite{li2021reinforcement}. 
Moreover, the command $c_t$ is also part of the observation. $c_t$ includes desired sagittal and lateral walking speed, walking height, and turning yaw rate ($c_t=[\dot{q}^d_x, \dot{q}^d_y, q^d_z, \dot{q}_{\phi}^d]$).

\paragraph{Action} The action $\mathbf{a}_t = q_m^d$ is the target position for the 10 actuated motors on Cassie which are first passed through a low-pass filter~\cite{li2021reinforcement} before being sent to the PD controller to generate the torque $u$ for each motor.

\subsection{Environment and Terrain variations}~\label{subsec:dynarand}
We randomize the environment parameters that includes robot modelling errors, sensor noise and delay, and terrain variations during the training. 
Based on previous work~\cite{li2021reinforcement}, the range of the randomization that is newly introduced is presented in Table.~\ref{tab:randomization}. 
Furthermore, we also introduce fractal-like terrain variations to make the policies robust.

\subsection{Training Details}
\paragraph{Base Policy and Environment Encoder} The base policy ($\pi$) is a MLP with 2 hidden layers of size $512$ each. The environment encoder ($\mu$) has 1 hidden layer of size $256$, and output ($z$) size of $8$. We train the policy with PPO using a batch size of $65536$ and minibatch size of $8192$.

\paragraph{Adaptation Module} The adaptation module ($\phi$) contains 3 convolution layers (kernels sizes $8, 5, 5$ and strides $4, 1, 1$ respectively) followed by 1 hidden layer of size 256. The output layer ($\hat{z}$) is $8$ dimensional. We train the policy with Adam optimizer using the same batch size, for a total of $2000$ iterations, resampling a new batch in every iteration. 

\paragraph{Base Policy Finetuning Stage} In the last finetuning stage of A-RMA, we freeze the adaptation module and finetune the base policy using PPO with a batch size of $65536$ and minibatch size of $8192$ for $2000$ iterations.

\section{Results and Analysis}
We first validate A-RMA in both MuJoCo and MATLAB Simulink (a high-fidelity simulator), and then present the results of RMA deployed on a bipedal robot Cassie in the real world. All A-RMA results in simulation and the real world are from the same policy which was only trained in MuJoCo and was deployed in MATLAB Simulink and the real world without any finetuning during test time. In this paper, we focus more on the control performance with a nominal walking height $q_z=0.98$~m, \textit{i.e.}, we do not change the walking height $q_z$ during the test though the policy is trained with variable walking height.

\begin{table}[t]
\begin{center}
\begin{tabular}{lccc}
\toprule
 & Min & Feasible & Tracking \\
 & Friction & Cmd Range & Error\\
 \midrule
 HZD Controller~\cite{li2020animated} & $0.3$ & $39/147$ & $[0.1554, 0.0811]$ \\
 Robust MLP~\cite{li2021reinforcement} & $0.3$ & $\mathbf{147/147}$ & $[0.0869, 0.1150]$\\
 RMA~\cite{rma} & $\mathbf{0.2}$ & $\mathbf{147/147}$ & $[0.0908, 0.0965]$\\
 A-RMA (Ours) & $\mathbf{0.2}$ & $\mathbf{147/147}$ & $\mathbf{[0.0849, 0.0952]}$\\
\bottomrule
\end{tabular}
\caption{\label{tab:benchmark1} \textbf{Comparison of A-RMA with baseline controllers in MATLAB Simulink.} We observe that all RL-controllers can track the entire command range. Of these, A-RMA and RMA generalize beyond the training friction range, maintaining stability in as low as a 0.2 friction coefficient. A-RMA has the best tracking performance in both $[\dot{q}^d_x, \dot{q}^d_y]$ compared to all the baseline controllers.}
\end{center}
\end{table}

\begin{table}[t]
\begin{center}
\begin{tabular}{lcccc}
\toprule
 & MTTF (s) & Return & Tracking Err & Mean Jerk\\ 
\midrule
A-RMA-static  & 7.1 & 158.5 & 0.79 & 0.64 \\
RMA~\cite{rma}  & 11.4 & 273.6 & 0.29 & 0.92 \\
A-RMA  & 14.0 & 335.8 & 0.30 & 0.30 \\
\midrule
A-RMA-priv  & 14.2 & 340.0 & 0.27 & 0.28 \\
\bottomrule
\end{tabular}
\caption{\label{tab:benchmark2} \textbf{Comparison of A-RMA with baseline controllers in MuJoCo.} We observe that A-RMA's performance is very close to A-RMA-priv, which has access to privileged simulation information, and is better than all the baseline controllers for the metrics: MTTF (max episode length = 20s), Returns and Mean Jerk experienced. The tracking performance (Mean Tracking Error) is similar to RMA. Note that RMA experiences a large drop compared to A-RMA-priv because of unobservability of the entire extrinsics vector from proprioception. A-RMA-static sees a sharp drop in performance compared to A-RMA, validating the importance of a continually and rapidly updating extrinsics vector.}
\end{center}
\vspace{-0.05in}
\end{table}

\begin{figure}[t]
\centering
\begin{subfigure}{.9\linewidth}
  \centering
  \includegraphics[width=\linewidth]{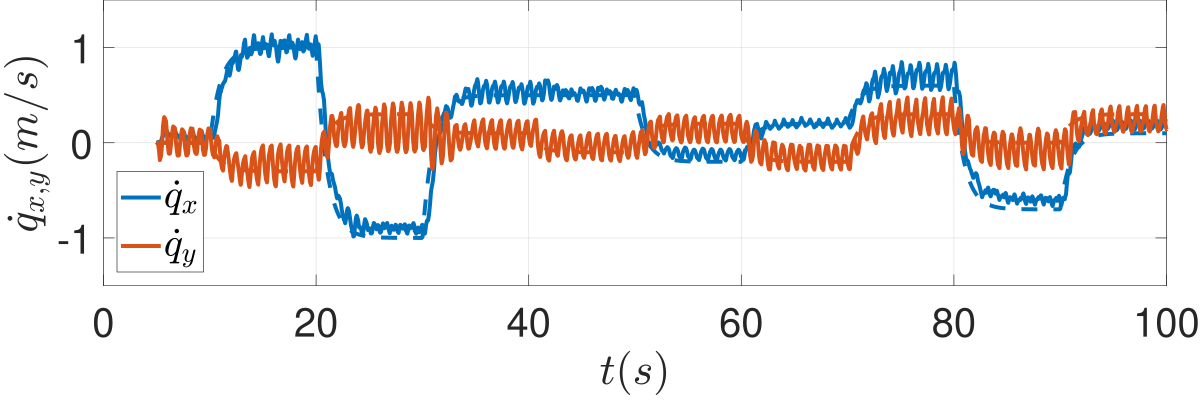}
  \caption{Robust MLP}
  \label{fig:robust-io}
\end{subfigure}
\begin{subfigure}{.9\linewidth}
  \centering
  \includegraphics[width=\linewidth]{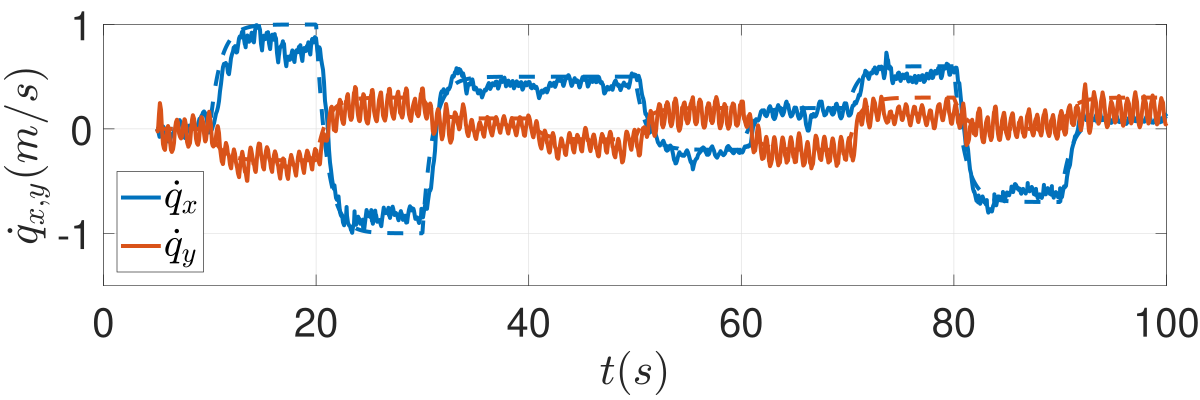}
  \caption{RMA}
  \label{fig:rma-dagger-io}
\end{subfigure}
\begin{subfigure}{.9\linewidth}
  \centering
  \includegraphics[width=\linewidth]{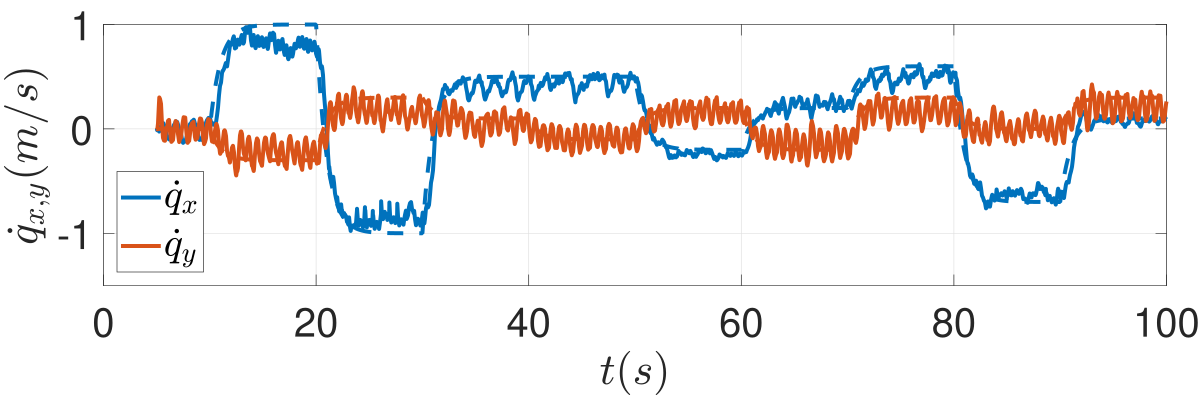}
  \caption{A-RMA}
  \label{fig:rma-ft-io}
\end{subfigure}
\caption{Plot of tracking error in sagittal and lateral directions different controllers. RMA and A-RMA produce less oscillations than the Robust MLP once the lateral velocity has stabilized.}\label{fig:tracking_error}
\vspace{-0.1in}
\end{figure}

\subsection{Simulation Validation}
The Simulink simulator provides us a safe validation environment to compare locomotion controllers while providing very high fidelity dynamics which is much more accurate than MuJoCo. We additionally show comparisons of A-RMA to its other adaptive variants in MuJoCo.  
Please note that the agent has no access to the Simulink data during the policy training in MuJoCo. 
We benchmark the A-RMA with the following locomotion controllers on Cassie as baselines: 1) a HZD-based controller (HZD) developed in~\cite{li2020animated} based on~\cite{gong2019feedback}, 2) a robust RL-based locomotion controller represented by MLP (Robust MLP) developed in~\cite{li2021reinforcement}, 3) A-RMA-static in which we estimate the latent in the first time step and then freeze it for the rest of the episode, 4) RMA~\cite{rma}. Finally as oracle, we also show the performance of A-RMA-priv which has access to privileged simulation information for reference.  
Please note that, for all the RL-based controllers, we utilize the same formulation of rewards and range of dynamics randomization presented in Sec.~\ref{subsec:dynarand} during training.

We benchmark these controllers using the following metrics: 1) the converged return, 2) command tracking performance on a nominal ground, 3) Minimum friction successfully handled, 4) Mean time to Fall (MTTF), 5) Mean Jerk of all the joints. In MuJoCo, the reported metrics are computed over 10 episodes with an episodic timeout of 20s during which we randomize all the parameters shows in Table~\ref{tab:randomization}. 

\paragraph{Converged Return and MTTF}
We report the performance of A-RMA-priv to understand the maximum achievable performance. As shown in Table~\ref{tab:benchmark2}, going from A-RMA-priv to RMA shows a sharp performance fall in both the metrics. This sharp fall is a consequence of non-negligible phase 2 regression loss as the extrinsics vector might not be fully observable from the proprioception history. We show that the proposed A-RMA recovers this performance drop and approximately matches the performance of the A-RMA-priv. We additionally see that A-RMA-static has a substantial performance fall indicating the importance of rapid online adaptation via estimated extrinsics.

\paragraph{Command Tracking}
We compare the different locomotion controllers on command tracking performance where we sample different desired sagittal walking speed $\dot{q}^d_x$ and lateral walking speed $\dot{q}^d_y$. 
The command tracking performance includes two parts: the range of the command for which a controller is able to maintain gait stability (Feasible Command Set~\cite{li2021reinforcement}), and tracking error between the desired and actual robot walking velocities.

\begin{figure*}[t]
\centering
\begin{subfigure}{.245\linewidth}
  \centering
  \includegraphics[width=\linewidth]{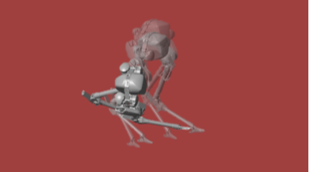}
  \caption{HZD}
  \label{fig:hzd-slip}
\end{subfigure}
\begin{subfigure}{.245\linewidth}
  \centering
  \includegraphics[width=\linewidth]{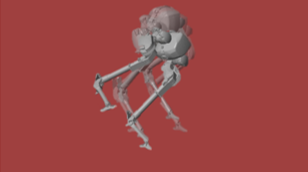}
  \caption{Robust MLP}
  \label{fig:robust-slip}
\end{subfigure}
\begin{subfigure}{.245\linewidth}
  \centering
  \includegraphics[width=\linewidth]{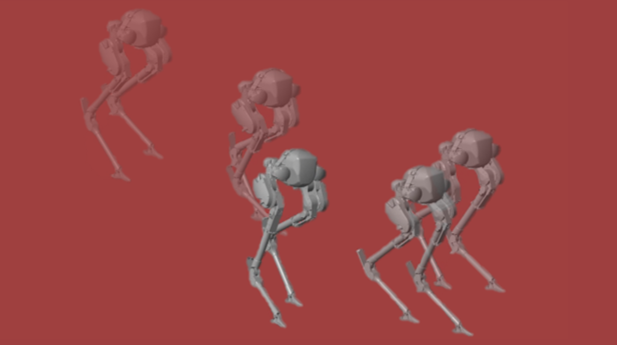}
  \caption{RMA}
  \label{fig:rma-dagger-slip}
\end{subfigure}
\begin{subfigure}{.245\linewidth}
  \centering
  \includegraphics[width=\linewidth]{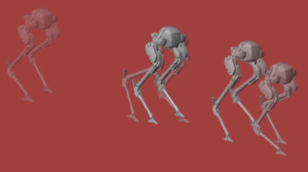}
  \caption{A-RMA}
  \label{fig:rma-ft-slip}
\end{subfigure}
\caption{Trajectories of Cassie on slippery ground with friction coefficient of $0.2$. We observe that A-RMA has minimal lateral deviation and is more stable than RMA, which shows significant deviation in the lateral walking direction.}\label{fig:slippery_sim}
\end{figure*}

\begin{figure*}[t]
\centering
\begin{subfigure}{.2125\linewidth}
  \centering
  \includegraphics[height=0.55\linewidth]{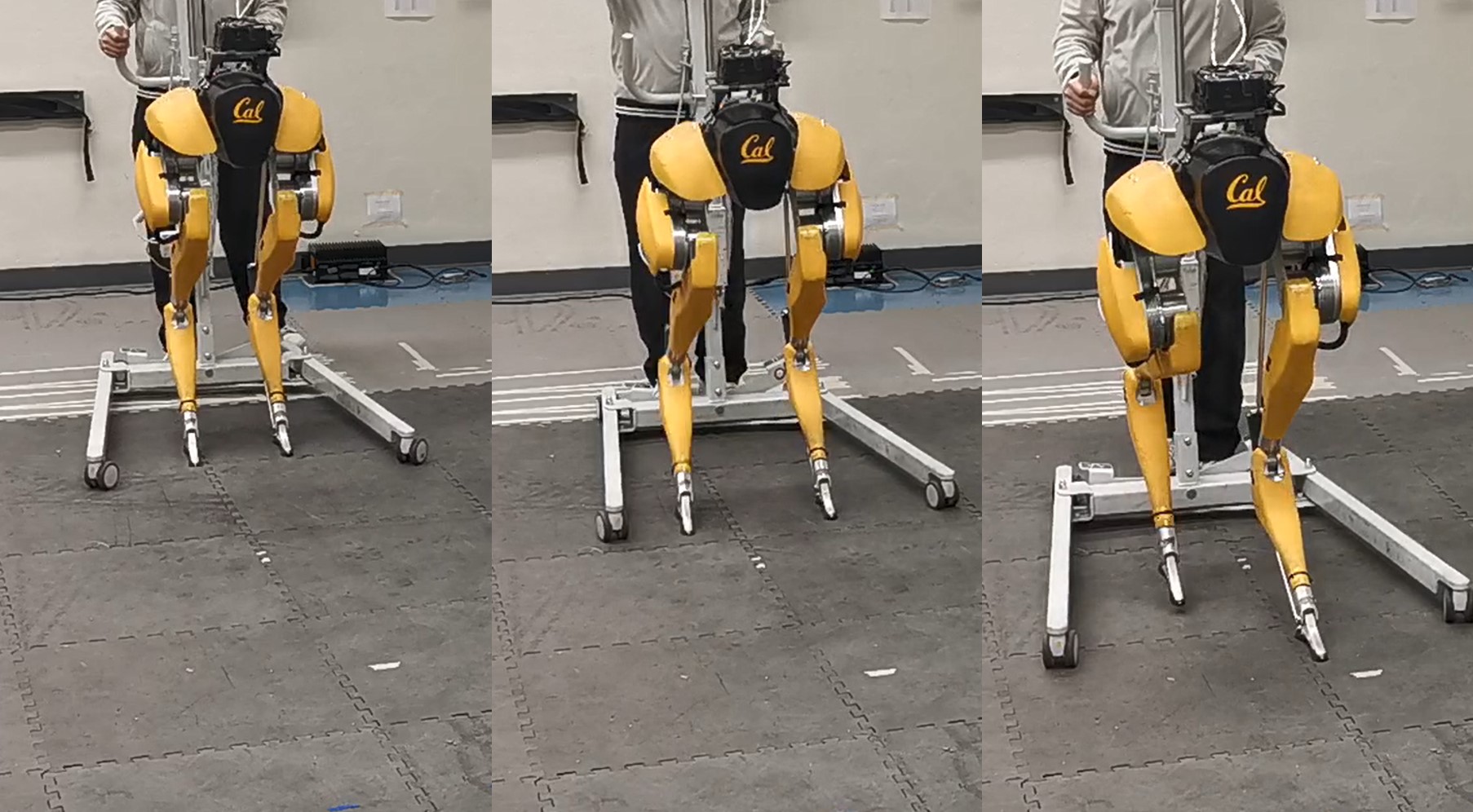}
  \caption{Nominal ground (forward)}
  \label{fig:flat-ground-forward}
\end{subfigure}
\hspace{0.3cm}
\begin{subfigure}{.2125\linewidth}
  \centering
  \includegraphics[height=0.55\linewidth]{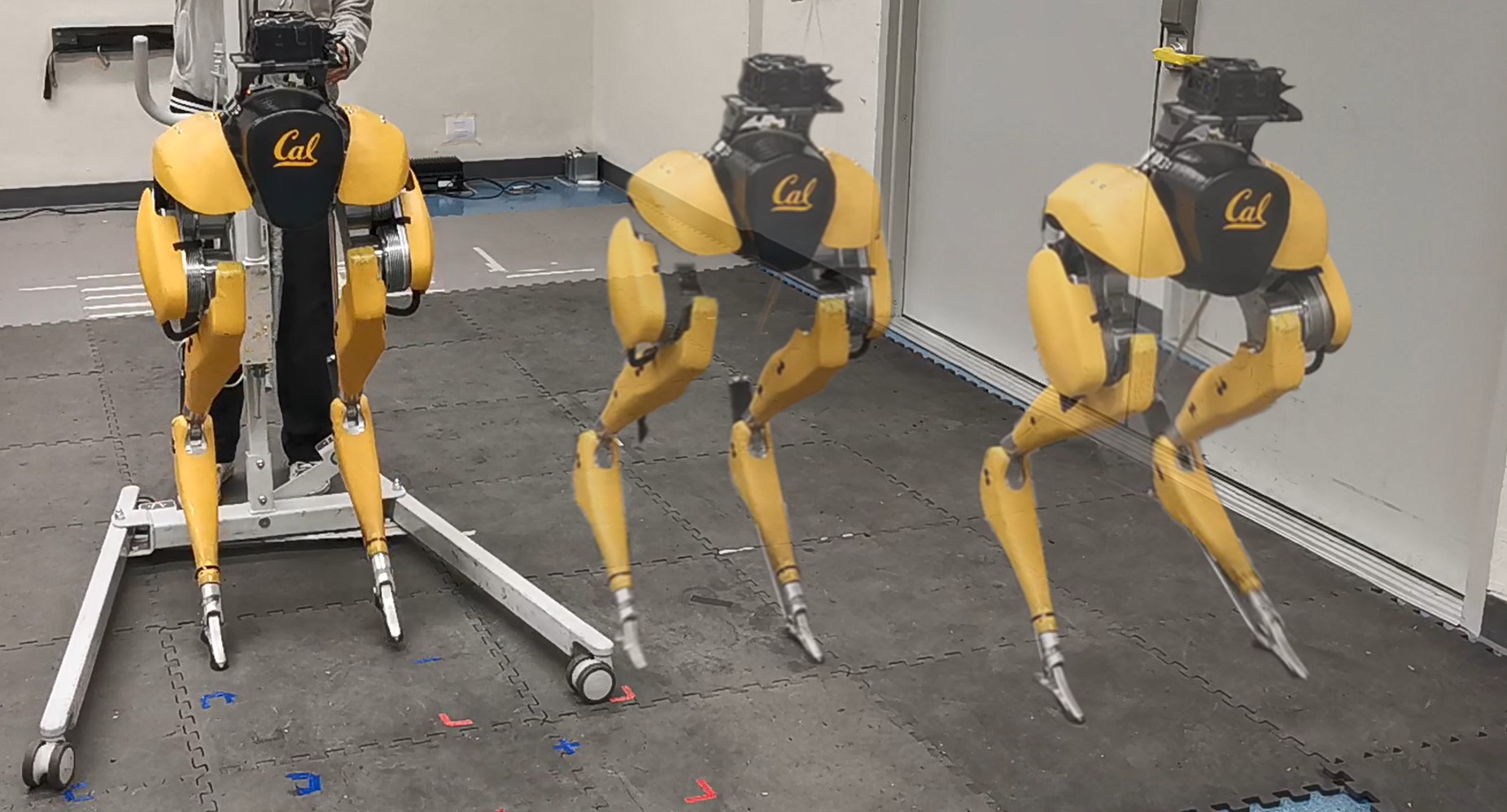}
  \caption{Nominal ground (sideways)}
  \label{fig:flat-ground-sidewalk}
\end{subfigure}
\begin{subfigure}{.535\linewidth}
  \centering
  \includegraphics[height=0.22\linewidth]{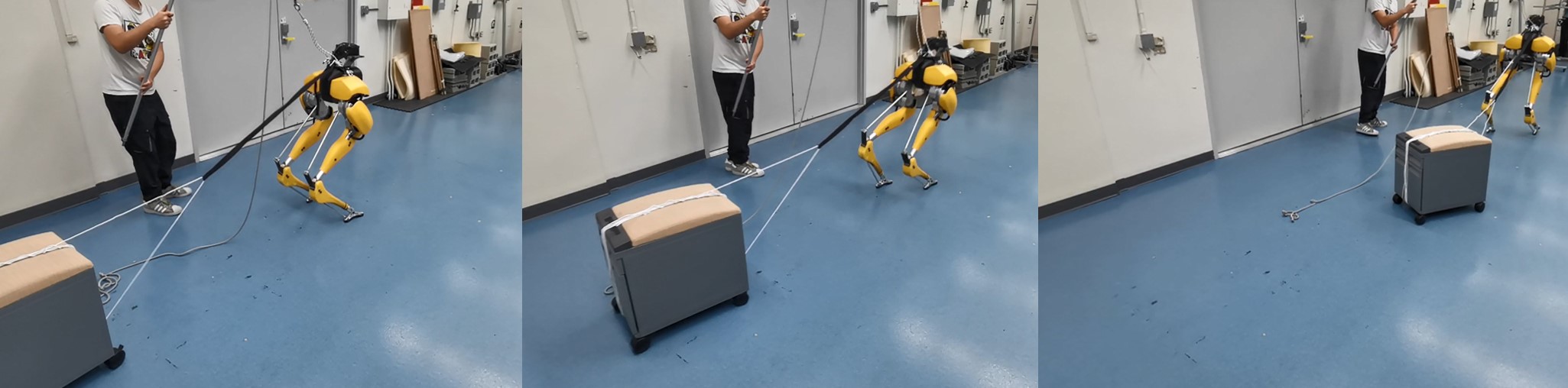}
  \caption{Towing a \raisebox{-0.5ex}{\textasciitilde}40 kg payload on wheels}
  \label{fig:load-lifting}
\end{subfigure}
\begin{subfigure}{.6525\linewidth}
  \centering
  \includegraphics[height=0.195\linewidth]{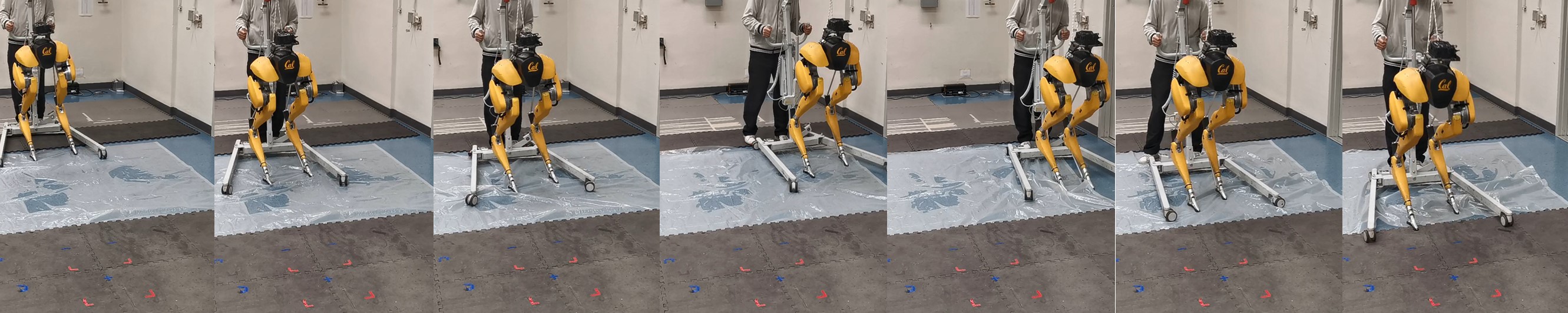}
  \caption{Slippery ground}
  \label{fig:slippery-ground}
\end{subfigure}
\begin{subfigure}{.32\linewidth}
  \centering
  \includegraphics[height=0.40\linewidth]{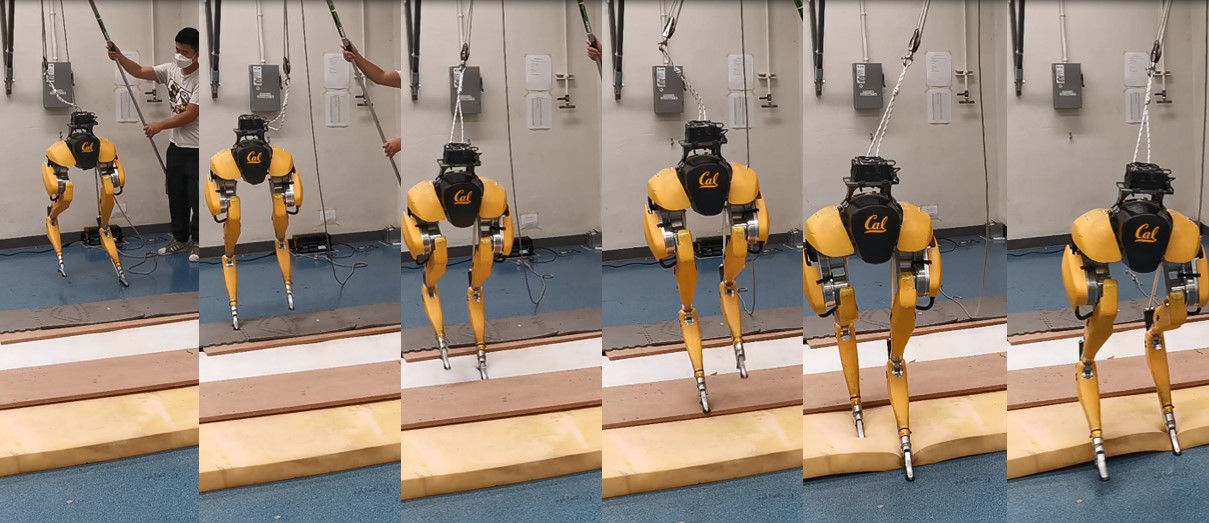}
  \caption{Uneven ground with variable softness}
  \label{fig:foam-ground}
\end{subfigure}
\caption{Real world deployment of Cassie walking on nominal ground, towing a payload, walking on a slippery surface, and on rough terrain with variable softness of the ground. The payload and rough terrain tests were repeated twice with consistent results. The same A-RMA policy was deployed in all these scenarios without any real world fine-tuning, task specific tuning, or calibration. We observe that A-RMA maintains stability in all of these deployment scenarios despite never having seen some of them during training.}\label{fig:experiments}
\vspace{-0.1cm}
\end{figure*}

We test the range of command $[\dot{q}^d_x, \dot{q}^d_y]^T$ from $[-1.0, -0.3]^T$ to $[1.0, 0.3]^T$ with a resolution of $[0.1, 0.1]^T$ with a nominal walking height $q_z = 0.98$ on Cassie with different locomotion controllers in simulation. 
These count for $147$ different commands and a command is considered feasible if the controller can maintain stability for $10$ seconds.
As demonstrated earlier in~\cite{li2021reinforcement}, HZD controller is only able to stably track a limited number of commands at nominal walking height. 
Specifically, the lateral walking speed cannot exceed $0.1~m/s$ while walking forwards, and there are only $39/147$ feasible commands using the HZD controller. 
However, all of the RL-based controllers can cover the entire range of the testing commands, \textit{i.e.}, there are $147/147$ feasible commands using locomotion controllers using Robust MLP, RMA, and A-RMA.
In terms of the tracking errors, as shown in Fig.~\ref{fig:tracking_error}, A-RMA manifests the least tracking errors with the only $[0.0849, 0.0952]^T~m/s$ compared to other RL-based controllers (Table.~\ref{tab:benchmark1} and Table.~\ref{tab:benchmark2}).
Moreover, as seen in Fig.~\ref{fig:tracking_error}, RMA and A-RMA produce less oscillations than Robust MLP. A-RMA has an overall superior performance in command tracking. 

\paragraph{Mean Jerk}
We measure the jerk experienced in the motors in MuJoCo for the baseline controllers. We observe in Table~\ref{tab:benchmark2} that A-RMA almost matches the smoothness of the A-RMA-priv, while RMA and A-RMA-static seem to do much worse. The additional phase 3 in A-RMA allows the base policy to learn smooth behaviours while accounting for the imperfect extrinsics estimator.   

\paragraph{Slippery Ground}
We use MATLAB Simulink to quantify the ability of locomotion controller to maintain gait stability on slippery ground. Fig.~\ref{fig:slippery_sim} shows that both HZD and Robust MLP fail to maintain robot balance when the ground friction coefficient is set to 0.2 while RMA and A-RMA succeed to control the robot to walk forwards and backwards.
We do a line search on friction values and observe that the minimal ground friction for which HZD and Robust MLP can maintain stability is 0.3 while RMA and A-RMA can go down to 0.2. 
This is significant because bipedal robots are inherently unstable and become very hard to control as the coefficient of friction approaches zero. 
Note that the RL-controllers were not trained for friction value of 0.2 (see Table~\ref{tab:randomization}), demonstrating that RMA and A-RMA have better generalization at test time than Robust MLP. 
Furthermore, as shown in Fig.~\ref{fig:rma-ft-slip}, Cassie does not deviate a lot in the lateral direction (with command $\dot{q}^d_y = 0$) when it walks backwards using A-RMA, while there exists a large drift to the right for RMA in Fig.~\ref{fig:rma-dagger-slip}.
This showcases A-RMA's superior tracking performance even on a slippery ground.

\subsection{Real World Experiments}

We deploy A-RMA on a Cassie bipedal robot in the real world, and test in four scenarios: 1) tracking variable commands on a nominal ground, 2) robot walking while towing a heavy load with cable, 3) slippery ground, and 4) rough terrain with variable softness.
We use the same policy for all our experiments which was trained in MuJoCo simulation and is deployed without any finetuning or calibration. The results are demonstrated in Fig.~\ref{fig:experiments}. Videos are available at~\cite{video} and included in the video attachment. 

As shown in Fig.~\ref{fig:flat-ground-forward}, \ref{fig:flat-ground-sidewalk}, Cassie is able to track varying commands including forwards, backwards, and sideways walking speed while maintaining gait stability and low ground impacts.

In the load carrying task, Cassie controlled by A-RMA is able to tow the load (around $40$ kg with wheels) while maintaining forward walking speed with no significant drift to other directions, as shown in Fig.~\ref{fig:load-lifting}.
During the tests, the robot is able to stay robust to the pulling force from the cable which fluctuates depending on the speed of the payload and the tension in the string making it hard to model.
Despite, slack/taut changes happening during the trials, the robot is able to maintain stability against such hybrid mode switches.

During the slippery ground test, we cover the ground by a plastic sheet with water in between in order to reduce the ground friction coefficient, as shown in Fig.~\ref{fig:slippery-ground}. 
When the robot steps onto the plastic sheet, there are significant slips between the robot feet and ground, \textit{i.e.}, unexpected contact changes which makes controlling a life-sized robot with such low ground traction very challenging.
However, the proposed policy is able to adapt to the changes in ground friction and contacts, and therefore able to maintain gait stability on such a slippery ground while tracking varying commands in both forward and lateral directions. 

We also introduce challenging types of terrains and contacts in the tests presented in Fig.~\ref{fig:foam-ground}, where we randomly place soft foams and wooden planks on the ground and let the robot walk onto that region. 
Such random changes of softness on the ground will change the contact type between the robot feet and ground: the contact region is on the sole when robot steps on a rigid plank while the contact can happens anywhere on the robot foot if it step into the soft foam, which making the contact very hard to model. 
Cassie, using A-RMA, is not only able to step on the rigid plank but also onto the soft foams without losing balance. 
Please note that the softness of the ground is not randomized during training as simulating a soft contact with high fidelity is still an open question. 
However, A-RMA is still able to generalize to the softness changes in the real world despite never having seen it during training.


\section{Conclusion} 
We study bipedal walking in complex, high-dimensional, life-sized Cassie robot. We propose A-RMA which extends the RMA algorithm by accounting for imperfect extrinsics estimation from the adaptation module. This is done by adding an additional base policy finetuning phase using model-free RL. This phase uses the imperfect extrinsics from adaptation module to allow the base policy to account for the unobservability in extrinsics. We have shown empirical gains in performance over model based as well as one of state-of-the art RL controllers. We train A-RMA on bipedal Cassie robot in simulation and then use the same A-RMA controller to experiment with walking on foam, slippery surface, rough terrain and payload towing in the real world. A-RMA shows generalization to terrains beyond what is seen during training without additional real-world finetuning or calibration. One limitation of the current work is that the robot is blind and only uses proprioception. For future work, it would be interesting to integrate on-board visual cameras for walking.

\section*{Acknowledgement}
This work was supported by the DARPA Machine Common Sense program and National Science Foundation Grant CMMI-1944722.

\bibliographystyle{IEEEtran}
\bibliography{main}
\balance

\end{document}